\documentclass[10pt,twocolumn,letterpaper]{article}

\usepackage{iccv}
\usepackage{times}
\usepackage{epsfig}
\usepackage{graphicx}
\usepackage{amsmath}
\usepackage{amssymb}
\usepackage{caption}
\usepackage{subcaption}

\usepackage[breaklinks=true,bookmarks=false]{hyperref}

\iccvfinalcopy % *** Uncomment this line for the final submission

\def\iccvPaperID{****} % *** Enter the ICCV Paper ID here

\ificcvfinal\fi

\begin{document}

%%%%%%%%% TITLE
\title{A New Dimension in Testimony: Relighting Video \\with Reflectance Field Exemplars}

\author{Loc Huynh \qquad Bipin Kishore \qquad Paul Debevec\\
USC Institute for Creative Technologies\\
{\tt\small \{lhuynh, kishore, debevec\}@ict.usc.edu}
}

% \maketitle
\twocolumn[{%
\renewcommand\twocolumn[1][]{#1}%
\maketitle
\begin{center}
    \centering
    \captionsetup{type=figure}
    \begin{subfigure}[b]{0.24\linewidth}
        \centering
        \captionsetup{type=figure}
        \includegraphics[width=\linewidth]{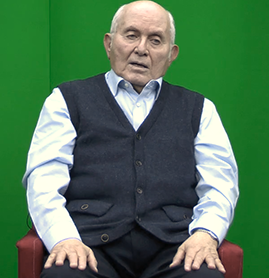}
        {(a) Input video}
    \end{subfigure}
    \begin{subfigure}[b]{0.75\linewidth}
        \centering
        \begin{subfigure}[b]{0.32\linewidth}
            \includegraphics[width=\linewidth]{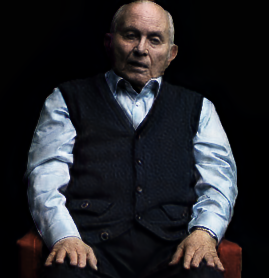}
        \end{subfigure}
        \begin{subfigure}[b]{0.32\linewidth}
            \includegraphics[width=\linewidth]{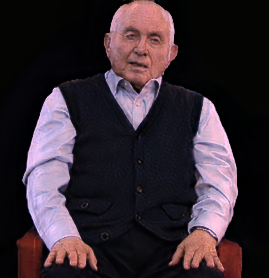}
        \end{subfigure}
        \begin{subfigure}[b]{0.32\linewidth}
            \includegraphics[width=\linewidth]{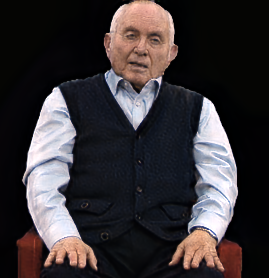}
        \end{subfigure}
        
        {(b) Subject relit with novel illuminations}
    \end{subfigure}
    \captionof{figure}{Given an input video of an actor lit by a single diffuse lighting condition (a), our method is able to relight the dynamic performance of the subject under any lighting condition (b).}
    \label{fig:teaser}
\end{center}
}]
% \begin{center}
%     \centering
%     \includegraphics[width=\linewidth]{figs/teaser_img.png}
%     \captionof{figure}{Given an input video of an actor lit by a single diffuse lighting condition (a), our method is able to relight the dynamic performance of the subject under any lighting condition (b).}
%     \label{fig:teaser}
% \end{center}
% }]

\ificcvfinal\thispagestyle{empty}\fi

%%%%%%%%% ABSTRACT
\begin{abstract}
\vspace{-0.1in}
We present a learning-based method for estimating 4D reflectance field of a person given video footage illuminated under a flat-lit environment of the same subject. For training data, we use one light at a time to illuminate the subject and capture the reflectance field data in a variety of poses and viewpoints. We estimate the lighting environment of the input video footage and use the subject's reflectance field to create synthetic images of the subject illuminated by the input lighting environment. We then train a deep convolutional neural network to regress the reflectance field from the synthetic images. We also use a differentiable renderer to provide feedback for the network by matching the relit images with the input video frames. This semi-supervised training scheme allows the neural network to handle unseen poses in the dataset as well as compensate for the lighting estimation error. We evaluate our method on the video footage of the real Holocaust survivors and show that our method outperforms the state-of-the-art methods in both realism and speed.
\vspace{-0.2in}
\end{abstract}

%%%%%%%%% BODY TEXT
\section{Introduction}
The New Dimensions in Testimony project at the University of Southern California's Institute for Creative Technologies recorded extensive question-and-answer interviews with twelve survivors of the World War II Holocaust.  Each twenty-hour interview, conducted over five days, produced over a thousand responses, providing the material for time-offset conversations through AI based matching of novel questions to recorded answers \cite{Artstein:2014:TOI}.  These interviews were recorded inside a large Light Stage system \cite{einarsson_relighting_2006} with fifty-four high-definition video cameras.  The multi-view data enabled the conversations to be projected three-dimensionally on an automultiscopic display \cite{Jones:2016:TOC,Jones:2015:APA}

The light stage system is designed for recording {\em relightable} reflectance fields, where the subject is illuminated from one lighting direction at a time, and these datasets can be recombined through image-based relighting \cite{Debevec2000}.  If the subject is recorded with a high speed video camera, a large number of lighting conditions can be recorded during a normal video frame duration \cite{Wenger2005,einarsson_relighting_2006} allowing a dynamic video to be lit with new lighting.  This enables the subject to be realistically composited into a new environment (for example, the place that the subject is speaking about) such that their lighting is consistent with that of the environment.  In 2012, the project performed a successful early experiment using a Spherical Harmonic Lighting Basis as in \cite{Fyffe:2011:CFP} for relighting a Holocaust survivor interview.  However, recording with an array of high speed cameras proved to be too expensive for the project, both in the cost of the hardware, and the greatly increased storage cost of numerous high-speed uncompressed video streams.

The project settled for recording the survivors in just a single interview lighting condition consisting of diffuse, symmetrical lighting from above.  But to enable relighting in the future, each survivor was recorded in a basis of forty-one lighting conditions in several static poses in a special session toward the end of each shoot as in Figure \ref{fig:olat}.  
The hope was that at some point, this set of static poses in different lighting conditions, plus the interview footage in diffuse lighting, could eventually be combined through machine learning to realistically show the interview as if it had been recorded in any combination of the lighting conditions, enabling general purpose relighting.  This paper presents a technique to achieve this goal, which provides a practical process for recording interview footage where the lighting can be controlled realistically after filming.

\begin{figure}[t]
\begin{center}
  \includegraphics[width=\linewidth]{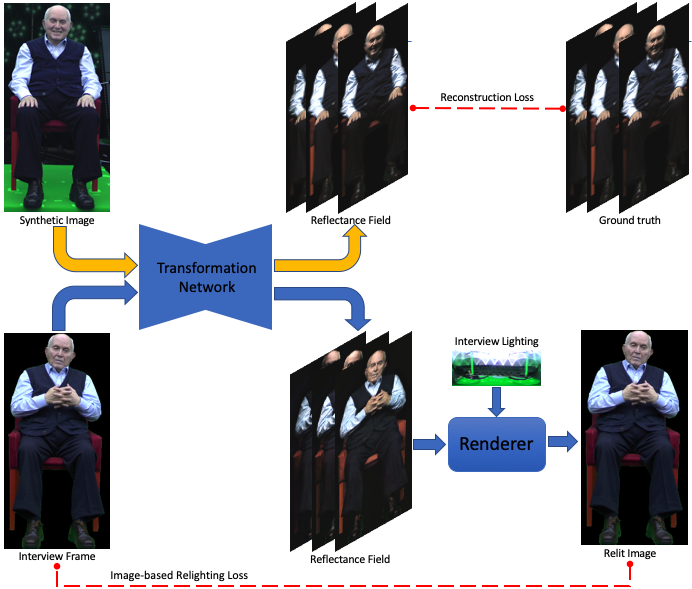}
\end{center}
  \caption{End-to-end semi-supervised training scheme. We use reconstruction loss for synthetic images while image-based lighting loss is applied to both real and synthetic interview images.}
\label{fig:training}
\end{figure}
% Data is important. Can adapt to later dataset.

%------------------------------------------------------------------------
\section{Related Work}

\begin{figure*}
\begin{center}
\includegraphics[width=\linewidth]{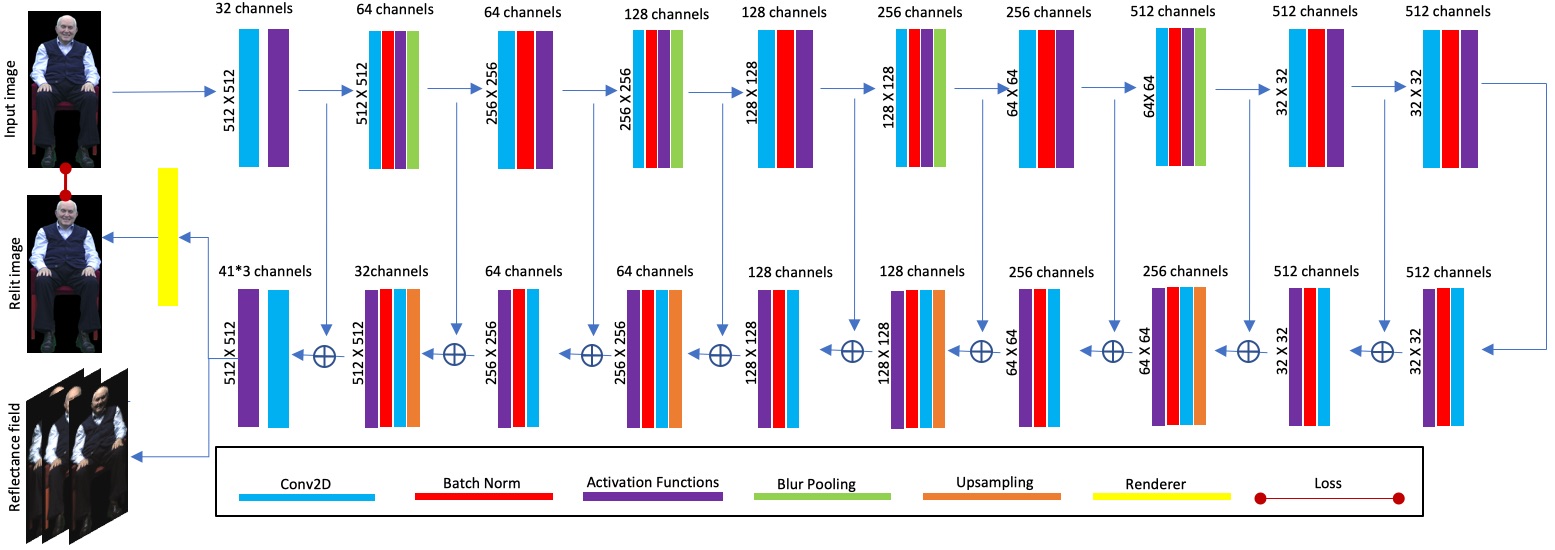}
\end{center}
  \caption{The architecture of our neural network. The input image is passed through a U-Net style architecture to regress to the set of OLAT images. When the ground truth is available, the network prioritizes the reconstruction loss of the OLAT imageset. Otherwise, the network is trained based on the feedback of the relit image.}
\label{fig:pipeline}
\end{figure*}

Relighting virtual humans from images and video is an active research topic in computer vision and computer graphics. In this section, we summarize some of the most related work in inverse rendering, image-based relighting, and learning-based relighting methods.

% % inverse rendering, monocular face and scene reconstruction methods, style transfer, photographic post-processing and deep learning.

% 0. LIghting capture
% 1. Model-fitting: classical graphic pipeline
% 2. Image-based: photographic post-processing
% 3. Learning-based: style transfer, deep reflectance field.

\emph{Inverse Rendering}. If photos of a scene can be analyzed to derive an accurate model of the scene's geometry and materials, the model can be rendered under arbitrary new lighting using forward rendering.  This {\em inverse rendering} problem is a long-studied research topic in computer vision and graphics \cite{Lombardi.Nishino2016,Ramamoorthi.Hanrahan2001,Yu99}. Many of these approaches use strong assumptions such as known illumination \cite{Horn1970}, or hand-crafted priors \cite{BarronTPAMI2015}. 

Unsurprisingly, relighting human bodies and faces has received particular interest recently. Many parametric models have been proposed to jointly reconstruct geometry, reflectance, and illumination of human bodies \cite{theobalt2007peoplelight}, faces \cite{Blanz.Vetter99,Garrido13,Garrido16,hawkins_animatable_2004,Ichim15,thies2016face}, eyes \cite{Berard16}, eyelids \cite{Bermano15}, and hair \cite{Hu15,Zhang17}.
\cite{li13} relights videos of humans based on estimation of parametric BRDF models and wavelet-based incident illumination.
\cite{Wen2003} uses a diffuse model for the face to relight it with a radiance environment map using ratio images.
\cite{Fyffe2009} performs relighting by using spherical gradient illumination images to fit a cosine lobe the reflectance function.
Several works estimate spatially-varying reflectance properties of a scene from either flash \cite{Nam18,li2018learning} or flat-lit images \cite{Gotardo18}.
\cite{Yamaguchi18} uses deep neural networks to estimate the parameters of a predefined geometry and reflectance model from a single image.

These parametric models are typically designed to handle specific parts of the human body. Many of these techniques rely on lightweight morphable models for geometry, a Lambertian model for skin reflectance, and a low-frequency 2\textsuperscript{nd} order spherical harmonic basis for illumination. Unfortunately, these strong priors only capture low-frequency detail and do not reproduce the appearance of specular reflections and sub-surface scattering in the skin.

In contrast, we use a deep neural network to infer the subject's reflectance field, which can be used to relight the images without explicitly modeling the geometry, material reflectance, and illumination of the images.

\emph{Image-based Relighting}. When a person is recorded under a large number of individual lighting conditions, they can be accurately relit by linearly combining those one-light-at-a-time (OLAT) images with the target illumination \cite{Debevec2000}. 
\cite{Wenger2005} used high-speed video to capture dynamic subjects with time-multiplexed lighting conditions for relighting, but required expensive cameras, optical flow computation, and was data intensive.  \cite{einarsson_relighting_2006} recorded a coarser lighting basis from a multitude of viewpoints, allowing post-production control of both the lighting and the viewpoint.  However, neither technique could be applied to long video segments due to the large size of the high frame rate video.
% Though effective in generating photo-realistic relighting, these techniques typically require a special setup from high-speed cameras, motionless subject to precise control over illuminations. 
% For a dataset of old people captured with only a few controlled lighting conditions such as Holocaust survivors dataset, these approaches are simply not applicable.

Another technique is to transfer reflectance field properties from a pre-captured subject to a target subject's performance as in \cite{Peers:2007:PFP}.  However, the quality of the lighting transfer depends on number of captured poses and the similarity in appearance of the two subjects.
Relighting can also be performed by transferring local image statistics from one portrait image to the target portrait as in \cite{Shih14}; however, this technique does not work well for extreme lighting changes.

Our approach also uses a set of lighting basis conditions to perform relighting. But instead of recording OLAT's for every moment of the video, our neural network infers OLATs for each video frame based on exemplars from static poses, enabling dynamic performance relighting.

\emph{Learning-Based Relighting}
\cite{Meka:2019} trains a deep neural network to map images of a subject lit by spherical gradient basis illumination to a set of one-light-at-a-time (OLAT) images for relighting.  Similar to this approach, we map the interview lighting images to a set of OLAT images. But unlike \cite{Meka:2019}, we employ a semi-supervised training scheme to train the network due to the lack of ground truth in our dataset and to work on the single interview lighting condition that's available. \cite{Tiancheng19} proposes a neural network that takes a single portrait photo under any lighting environment and relights the subject with arbitrary target illumination. This network was trained on a large set of subjects individuals under a dense set of lighting conditions to predict the input illumination and perform relighting by replacing the illumination at the bottleneck of the neural network. The technique is overall successful, but the low resolution of the predicted illumination limits the quality of the relit result.  \cite{LeGendre2019} also uses multitudes of data from 70 subjects with dense lighting conditions to estimate a more detailed HDR lighting environment from a single portrait image.  \cite{Texler20} presents a recent advance in Style Transfer techniques, where a video can be changed to a different style by registering to one or more keyframes in the new style.  This is most often used to transfer non-photorealistic rendering styles such as a pastel drawing, but can also be used to transfer a new style of lighting.  However, this technique has not been applied to create arbitrarily relightable models and requires registration from the style exemplars to the video sequence.  In comparison, our method is designed to perform realistic relighting from a single lighting condition by providing the neural network a set of reflectance field exemplars of how the subject actually should appear under OLAT lighting conditions.

\section{Method}
One of the most effective ways to perform realistic relighting is to combine a dense set of basis lighting conditions (a \emph{reflectance field}) with according to a novel lighting environment to simulate the appearance in the new lighting.  However, this approach is not ideal for a dynamic performance since it requires either high-speed cameras, or requires the actor to sit still for several seconds to capture the set of OLAT images. \cite{Meka:2019} overcomes this limitation by using neural networks to regress 4D reflectance fields from just two images of a subject lit by gradient illumination. They postulate that one can also use flat-lit images to achieve similar results with less high-frequency detail. Since the method casts relighting as a supervision regression problem, it requires pairs of tracking images and their corresponding OLAT images as ground truth for training.

In the New Dimensions in Testimony project, most of the Holocaust survivors' interview footage was captured in front of a green screen so that the virtual backgrounds can be added during post-production. However, this setup poses difficulties for achieving consistent illumination between the actors and the backgrounds in the final testimony videos and does not provide the ground truth needed for supervision training.  In this paper, we use the limited OLAT data to train a neural network to infer reflectance fields from synthetically relit images. The synthetic relit images are improved by matching them with the input interview images through a differentiable renderer, enabling an end-to-end training scheme. For more training details, see Figure \ref{fig:training}.
% One of the most effective ways to perform realistic relighting is to combine a finite set of lighting conditions (also known as reflectance field) with the environment map to simulate the lighting in that map. However, this approach is not ideal for dynamic performance since it requires the actor to sit still for several seconds to capture one set of OLAT images.
% The goal of our work is photo-realistically relight the interview footage of the actors by directly predicting the reflectance field for every single frame of the footage.

In this section, we describe the data acquisition process, how we relate the OLAT reflectance field exemplars with the interview footage, and how we train an end-to-end neural network to regress reflectance fields for realistic relighting.

\subsection{Data Acquisition and Processing}
\begin{figure}[t]
\begin{center}
  \includegraphics[width=\linewidth]{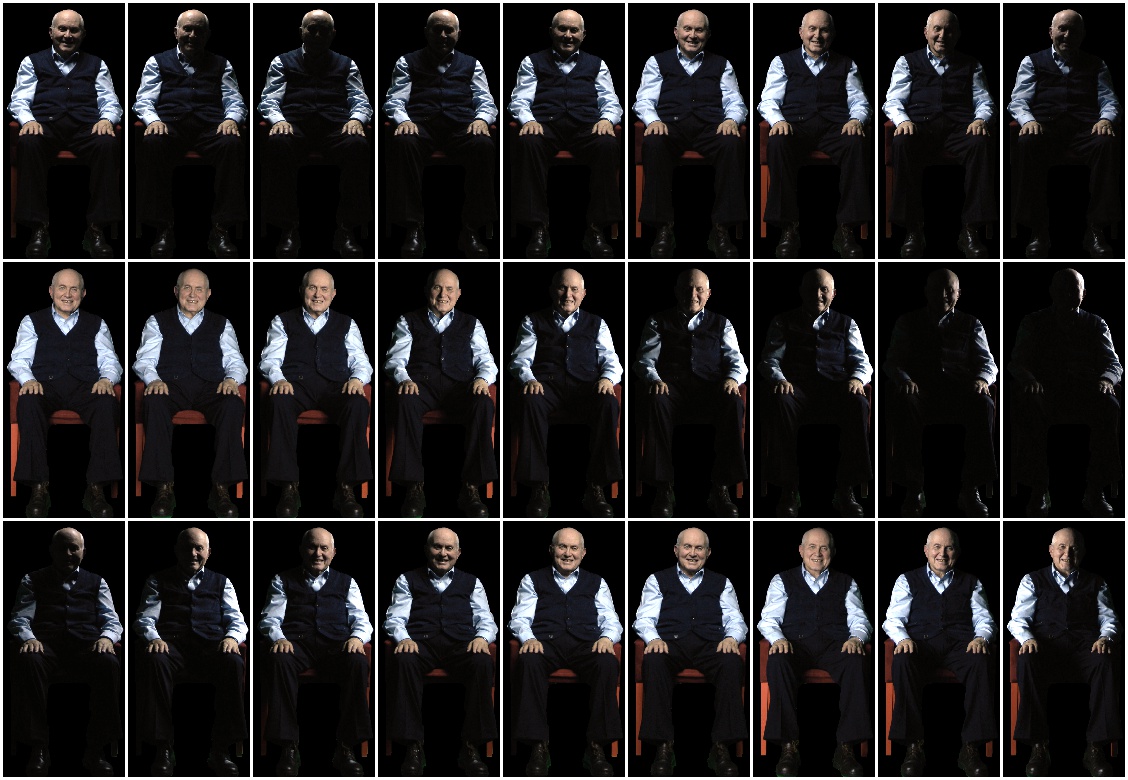}
\end{center}
  \caption{Reflectance field: 27 of 41 one-light-at-a-time images.}
\label{fig:olat}
\end{figure}
Each Holocaust survivor was recorded over a 180-degree field of view using an array of 50 Panasonic X900MK 60fps progressive scan consumer camcorders, each four meters away framed on the subject.
Toward the end of each Holocaust survivor's lengthy interview, they were captured in several different static poses under a reflectance field lighting basis of 41 lighting conditions as in Figure \ref{fig:olat}.  The lighting conditions were formed using banks of approximately 22 lights each of the 931 light sources on the 8m diameter dome \cite{einarsson_relighting_2006}.  This somewhat lower lighting resolution was chosen to keep the capture time shorter than what would be required to record each of the 931 lights individually and to avoid too great of a degree of underexposure so that we could use the same exposure settings as the interview lighting without touching the cameras. %without too great of a degree of underexposure.

% For each OLAT capture session, the actor performs six to ten poses, and we acquire a sequence of photographs capturing an actor under 41 different illumination patterns (see ). The illumination patterns are produced by a programmable illumination device, with 901 controllable light sources arranged in a sphere, similar to . Unlike the lighting basis design of \cite{einarsson_relighting_2006}, we use triples of lights as if they were single individual lighting directions to increase illumination for each OLAT image. Though the resolution of the target lighting environments is reduced by the lack of lighting basis, we observe that the effect is not particularly noticeable.

\textit{Data Processing.}
The original resolution of our video frames images is $1920 \times 1080$ in portrait orientation. For each image, we crop the full body of the actors, and then use Grabcut \cite{Rother2004} to mask out the background. The images are then padded and resized to  $512 \times 512$.

\textit{Synthetic Tracking Frames.} 
We use a mirror ball image captured right after the interview session as a light probe \cite{Debevec98}. This light probe represents the illumination of the interview session. For convenience, we convert the light probe to a latitude-longitude format.
Then we use mirror ball images captured in the OLAT session to find their projections in our target environment illumination map. By taking a weighted combination of the images in the OLAT set according to these projections, we are able to relight all the static poses of the actor.

The OLAT images are not as well exposed as the interview lighting since fewer light sources are on, and we discovered that the consumer video cameras applied a weaker level of gamma correction to the darker range of pixel values, making dark regions appear even darker, presumably as a form of noise suppression.  Thus, we developed a dual gamma correction curve to linearize the image data:
\begin{equation}
I'=(1-I)*I^{\gamma_1}+I*I^{\gamma_2}
\end{equation}
where $\gamma_1$, $\gamma_2$ describe the gamma we use for the lower and upper part of the gamma curve, and we interpolate between these two curves according to the brightness of the pixel. We optimize $\gamma_1$, $\gamma_2$ so that the OLAT reflectance field exemplars, relit with the measured interview lighting condition, match the appearance of the first frame of the interview video. Though each subject is only recorded as a reflectance field in a few poses, these synthetic relit images play an important role in bringing the output of the network closer to the illumination of the input video footage.

\subsection{Network Architecture}
We cast the relighting problem as prediction of the reflectance field, and use these measurements to render the subject under arbitrary illumination. To be consistent with the Holocaust survivor dataset, we define a reflectance field to have 41 OLAT images. Our goal is to predict how the actor would look under 41 specified lighting conditions for every frame of dynamic performance. The structure of our neural network resembles the structure of the popular image transformation architecture with skip connections \cite{Ronneberger15}. 
The encoder consists of ten blocks of $3\times3$ convolution layers each followed by a batch-normalization layer and a leaky ReLU activation function. A blur-pooling operation \cite{zhang2019shiftinvar} is used at the end of the block is to decrease the spatial resolution and increase the number of channels. Note that the first block of the encoder does not have a batch-normalization layer and uses a $7\times7$ convolution layer.

The decoder follows a similar structure with ten blocks of bilinear upsampling followed by a convolution layer. At the end of each decoder block, we use skip connections to concatenate the network features with their corresponding activations in the encoder. All convolution layers are followed by a ReLU activation except for the last convolution layer where a sigmoid activation is used. At the end of the decoder is a differentiable renderer that takes as input a whole set of OLAT images to render a subject under a new calibrated illumination condition. For network details, see Figure \ref{fig:pipeline}.

\begin{figure}[t]
     \centering
     \begin{subfigure}[b]{0.32\linewidth}
         \centering
         \includegraphics[width=\linewidth]{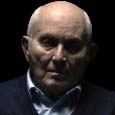}
         \caption{Reference}
     \end{subfigure}
     \begin{subfigure}[b]{0.32\linewidth}
         \centering
         \includegraphics[width=\linewidth]{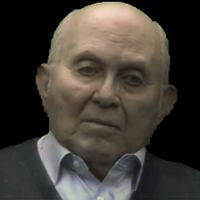}
         \caption{Sun et al.\cite{Tiancheng19}}
     \end{subfigure}
     \begin{subfigure}[b]{0.32\linewidth}
         \centering
         \includegraphics[width=\linewidth]{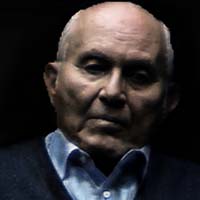}
         \caption{Ours}
     \end{subfigure}
        \caption{Comparison with Single Image Portrait Relighting. Our result has greater lighting detail and looks much closer to the reference lighting.}
        \label{fig:vsPortrait}
\end{figure}

\subsection{Loss Function}
Our model is trained through the minimization of a weighted combination of two loss functions. A reconstruction loss minimizes errors between the set of OLAT images in the dataset and set of OLAT images predicted by the network. The second loss is an image-based relighting loss that minimizes the errors between the input image and the rendered image lit with the predicted reflectance field. The backgrounds are masked out in all loss calculations.

\textit{Reconstruction Loss.} This loss ensures the accurate inference of the network by matching the network prediction with the ground truth. Since the per-pixel photometric loss often leads to blurry output images, we choose to minimize the loss in feature space with a perceptual loss. Letting \( VGG^{(i)}(I) \) be the activations of the $i$th layer of a VGG network \cite{Simonyan15}, the reconstruction loss is defined as:
\begin{equation}
L_{rec} = \sum_{i=1}^N \sum_{j=1}^{M} \|VGG^{(j)}(I_{pred}^i) - VGG^{(j)}(I_{gt}^i)\|_2
\end{equation}
where $N$ is the number of images in a complete OLAT set, and $M$ is the number of VGG layers to be used.

\textit{Image-Based Relighting Rendering Loss} This self-supervision loss makes the network more robust to unseen poses of the actor in the training set. Given a predicted reflectance field \(R(\theta,\phi,x,y)\) and the calibrated interview lighting environment $L_i$, we can relight the actor as follows:
\begin{equation}
I_{relit} = \sum_{\theta,\phi}R_{x,y}(\theta,\phi)L_i(\theta,\phi)
\end{equation}
where \(R_{x,y}(\theta,\phi)\) represents how much light is reflected toward the camera by pixel \((x,y)\) as a result of illumination from direction \((\theta,\phi)\). Matching this relit image $I_{relit}$ and the input image $I$ gives us the rendering loss:
\begin{equation}
L_{render} = \sum_{j=1}^{M} \|VGG^{(j)}(I_{relit}) - VGG^{(j)}(I)\|_2
\end{equation}
The full objective is the weighted combination of the two loss functions:
\begin{equation}
L = \lambda_1L_{rec} + \lambda_2L_{render}
\end{equation}

\textit{Implementation details.}
We use two sets of data to train the network. The first set consists of six poses with groundtruth OLAT images and the six corresponding relit images showing the reflectance field exemplars under the simulated interview lighting condition. The second set consists of 100 frames of the target video.
We train the first set with both a reconstruction loss $L_{rec}$ and a rendering loss $L_{render}$ for 100 epochs, and then we train the second set for 4 epochs with only the rendering loss before going back to supervised training. The training process continues until we reach 1040 epochs. We use the ADAM optimizer \cite{KingmaB14} with \(\beta_1=0.9, \beta_2=0.999\) and a learning rate of 0.001.

\section{Evaluation}

\begin{figure}[t]
     \centering
     \begin{subfigure}[b]{0.32\linewidth}
         \centering
         \includegraphics[width=\linewidth]{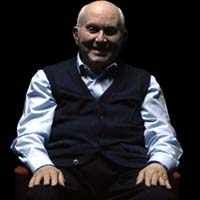}
         \caption{Reference}
     \end{subfigure}
    %  \hfill
     \begin{subfigure}[b]{0.32\linewidth}
         \centering
         \includegraphics[width=\linewidth]{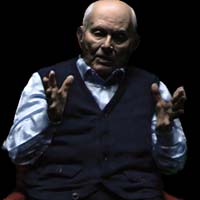}
         \caption{Texler et al. \cite{Texler20-CAG}}
     \end{subfigure}
    %  \hfill
     \begin{subfigure}[b]{0.32\linewidth}
         \centering
         \includegraphics[width=\linewidth]{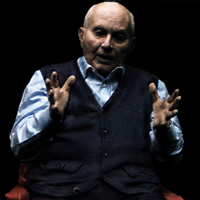}
         \caption{Ours}
     \end{subfigure}
        \caption{Comparison with Style Transfer based relighting. Our method reproduces more convincing  shadows and highlights.}
        \label{fig:style}
\end{figure}

\begin{figure*}
\centering
    \begin{subfigure}[t]{\dimexpr0.105\linewidth+20pt\relax}
        \makebox[20pt]{\raisebox{40pt}{\rotatebox[origin=c]{90}{Input Video}}}%
        \includegraphics[width=\dimexpr\linewidth-20pt\relax]{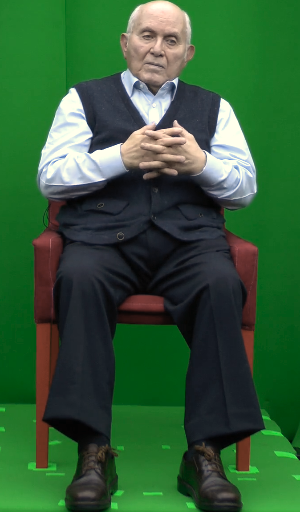}
        \makebox[20pt]{\raisebox{40pt}{\rotatebox[origin=c]{90}{OLAT 1}}}%
        \includegraphics[width=\dimexpr\linewidth-20pt\relax]{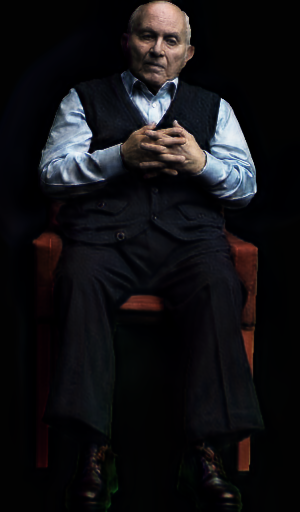}
        \makebox[20pt]{\raisebox{40pt}{\rotatebox[origin=c]{90}{OLAT 2}}}%
        \includegraphics[width=\dimexpr\linewidth-20pt\relax]{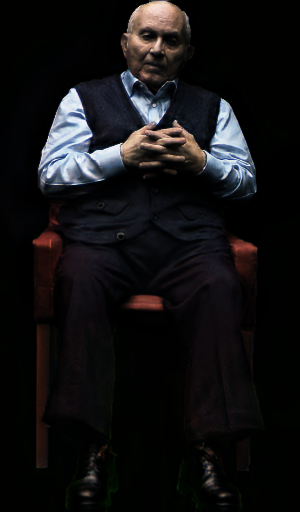}
        \makebox[20pt]{\raisebox{40pt}{\rotatebox[origin=c]{90}{Grace Cathedral}}}%
        \includegraphics[width=\dimexpr\linewidth-20pt\relax]{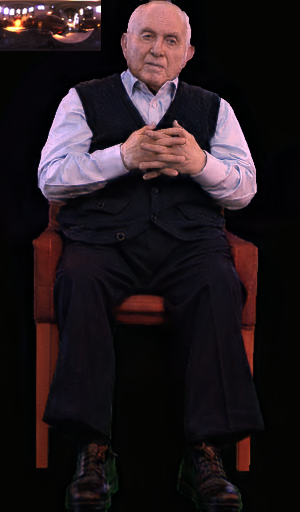}
        \makebox[20pt]{\raisebox{40pt}{\rotatebox[origin=c]{90}{Pisa Courtyard}}}%
        \includegraphics[width=\dimexpr\linewidth-20pt\relax]{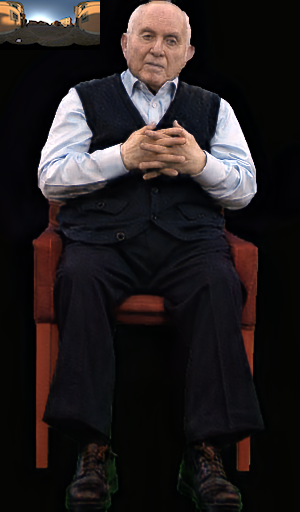}
        % \caption{model A}
    \end{subfigure}\hfill
    \begin{subfigure}[t]{0.105\linewidth}
        \includegraphics[width=\linewidth]{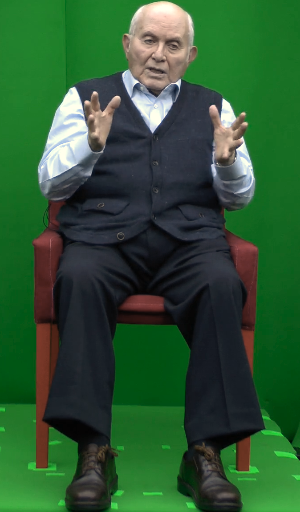}
        \includegraphics[width=\linewidth]{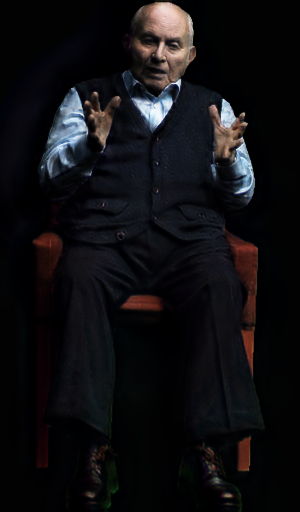}
        \includegraphics[width=\linewidth]{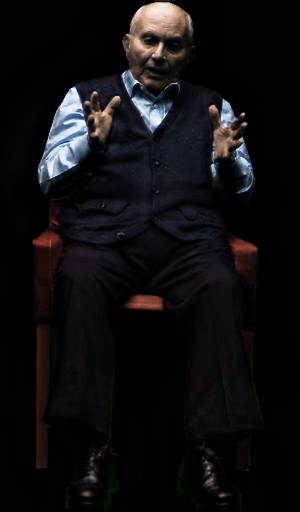}
        \includegraphics[width=\linewidth]{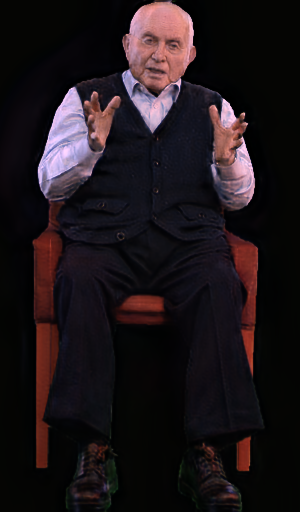}
        \includegraphics[width=\linewidth]{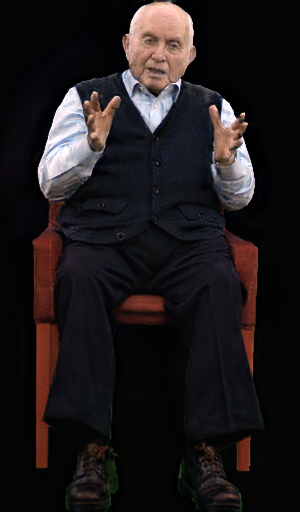}
        % \caption{model A}
    \end{subfigure}\hfill
    \begin{subfigure}[t]{0.105\linewidth}
        \includegraphics[width=\linewidth]{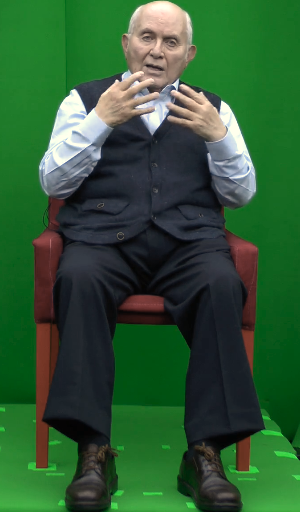}
        \includegraphics[width=\linewidth]{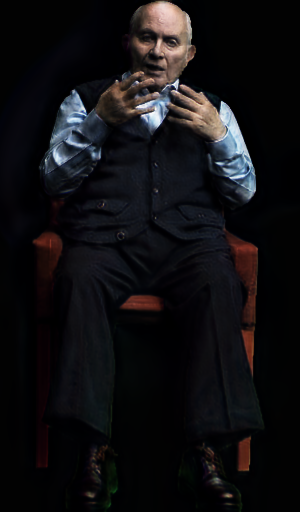}
        \includegraphics[width=\linewidth]{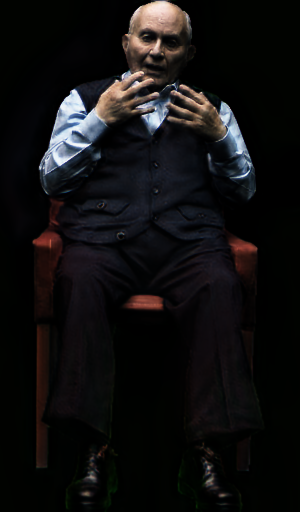}
        \includegraphics[width=\linewidth]{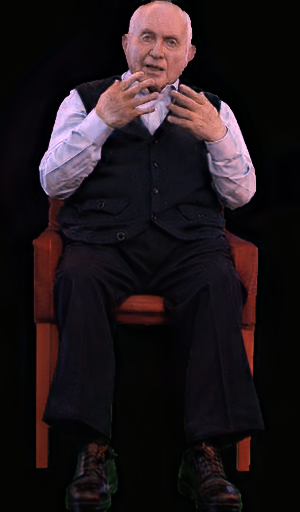}
        \includegraphics[width=\linewidth]{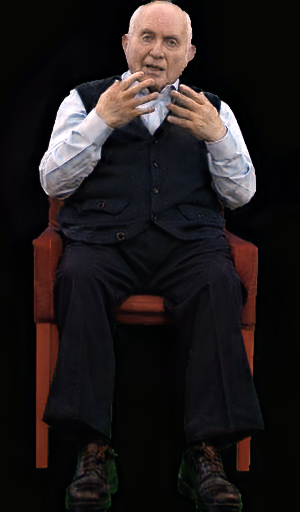}
        % \caption{model B}
    \end{subfigure}\hfill
    \begin{subfigure}[t]{0.105\linewidth}
        \includegraphics[width=\linewidth]{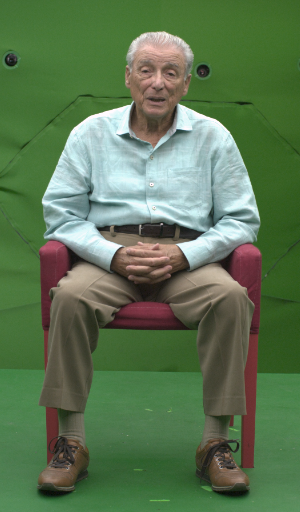}
        \includegraphics[width=\linewidth]{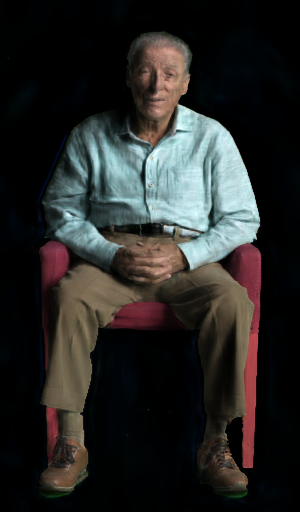}
        \includegraphics[width=\linewidth]{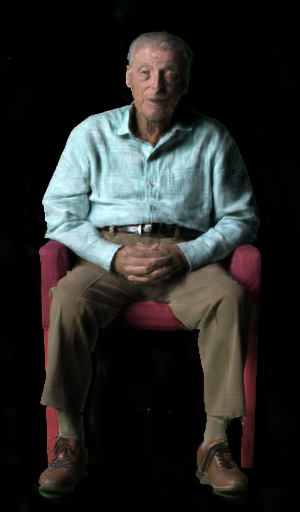}
        \includegraphics[width=\linewidth]{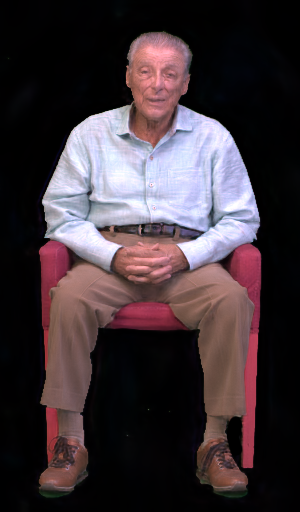}
        \includegraphics[width=\linewidth]{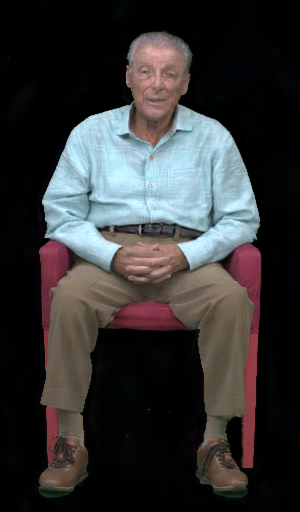}
        % \caption{model B}
    \end{subfigure}\hfill
    \begin{subfigure}[t]{0.105\linewidth}
        \includegraphics[width=\linewidth]{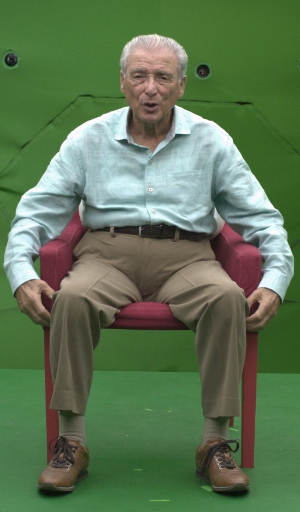}
        \includegraphics[width=\linewidth]{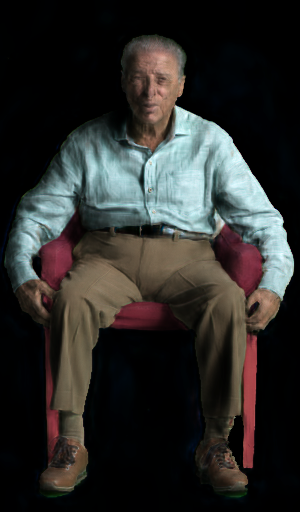}
        \includegraphics[width=\linewidth]{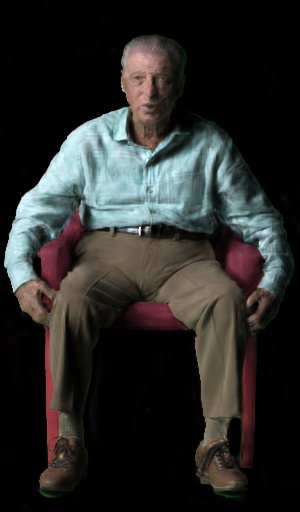}
        \includegraphics[width=\linewidth]{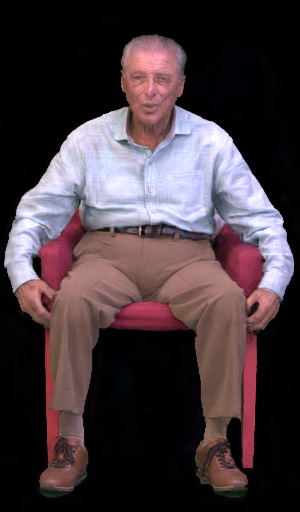}
        \includegraphics[width=\linewidth]{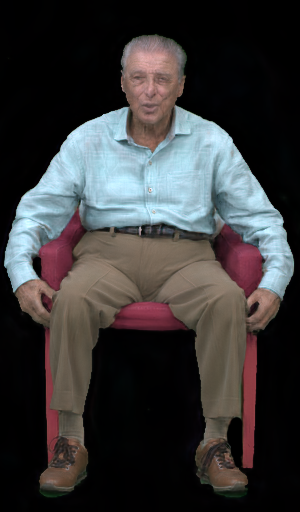}
        % \caption{model B}
    \end{subfigure}\hfill
    \begin{subfigure}[t]{0.105\linewidth}
        \includegraphics[width=\linewidth]{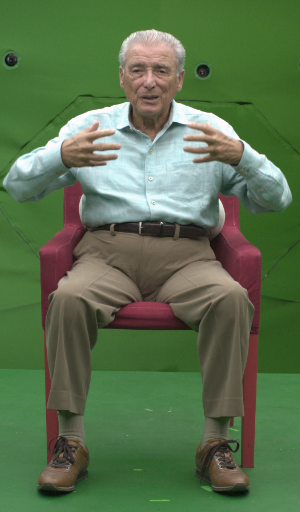}
        \includegraphics[width=\linewidth]{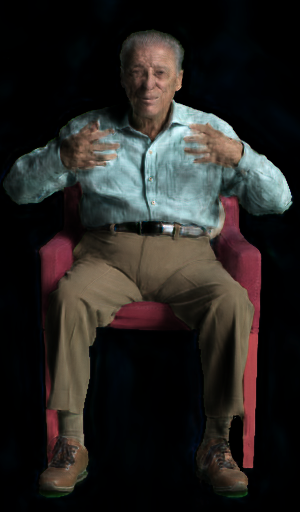}
        \includegraphics[width=\linewidth]{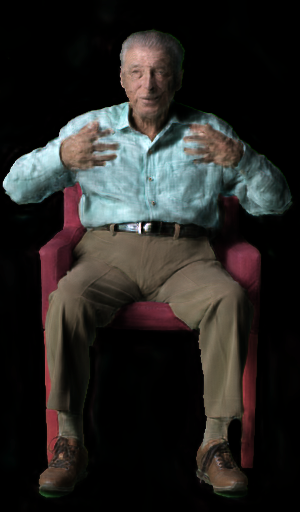}
        \includegraphics[width=\linewidth]{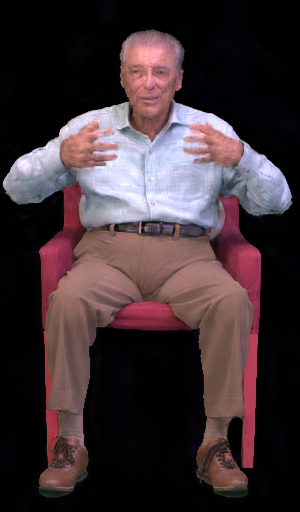}
        \includegraphics[width=\linewidth]{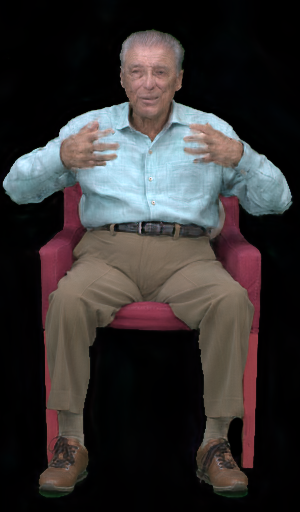}
        % \caption{model B}
    \end{subfigure}\hfill
    \begin{subfigure}[t]{0.105\linewidth}
        \includegraphics[width=\linewidth]{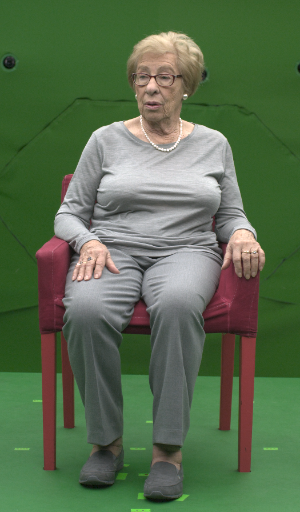}
        \includegraphics[width=\linewidth]{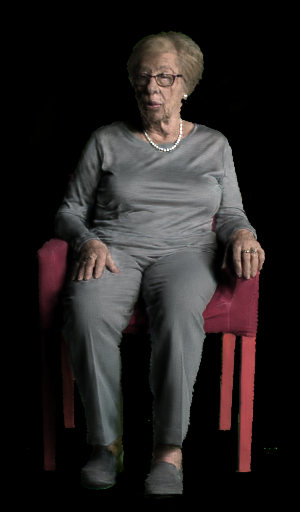}
        \includegraphics[width=\linewidth]{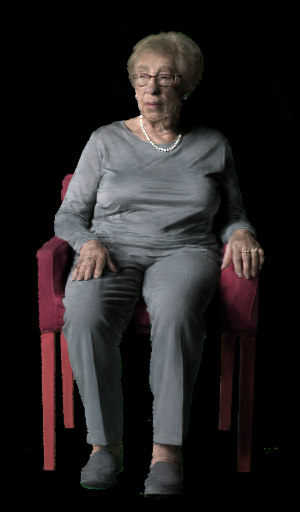}
        \includegraphics[width=\linewidth]{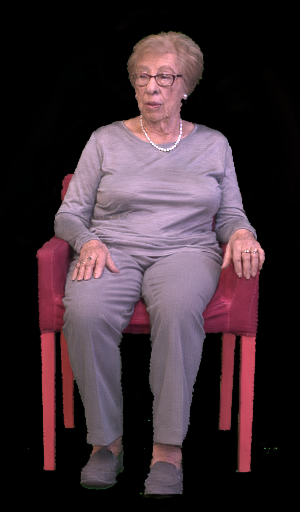}
        \includegraphics[width=\linewidth]{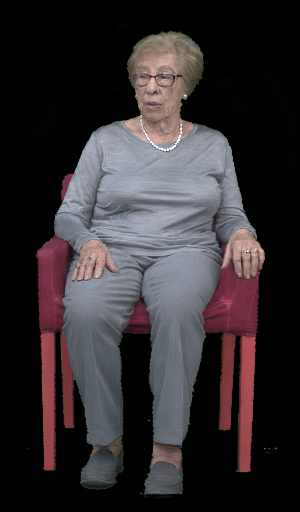}
        % \caption{model B}
    \end{subfigure}\hfill
    \begin{subfigure}[t]{0.105\linewidth}
        \includegraphics[width=\linewidth]{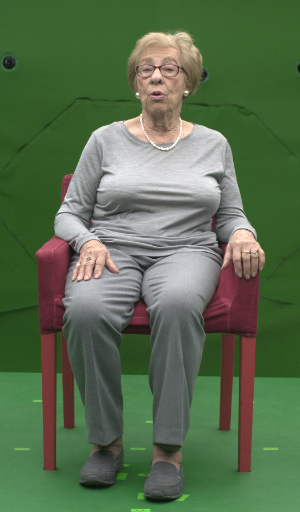}
        \includegraphics[width=\linewidth]{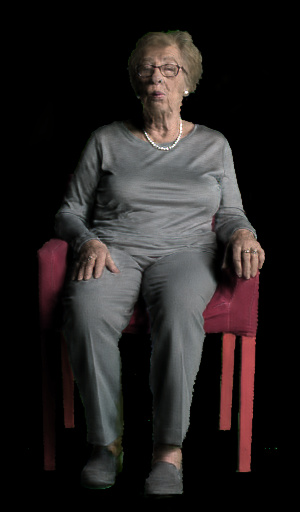}
        \includegraphics[width=\linewidth]{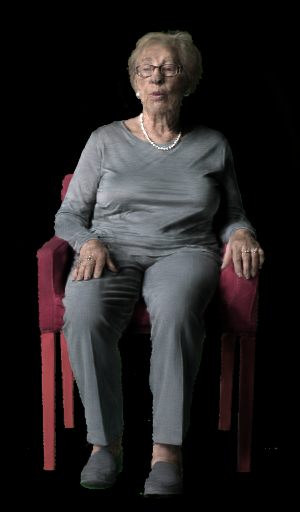}
        \includegraphics[width=\linewidth]{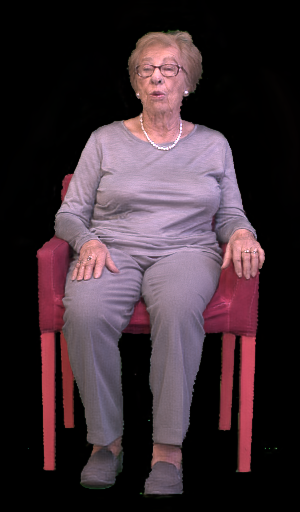}
        \includegraphics[width=\linewidth]{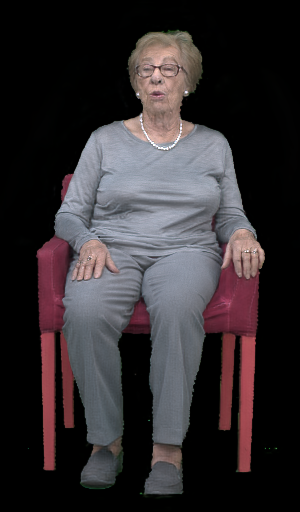}
        % \caption{model B}
    \end{subfigure}\hfill
    \begin{subfigure}[t]{0.105\linewidth}
        \includegraphics[width=\linewidth]{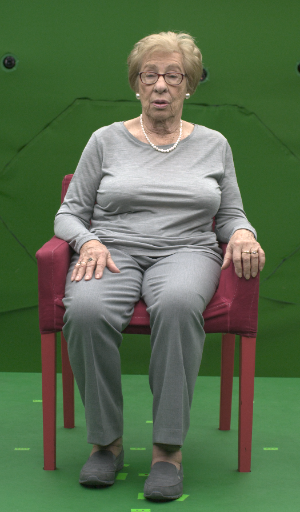}
        \includegraphics[width=\linewidth]{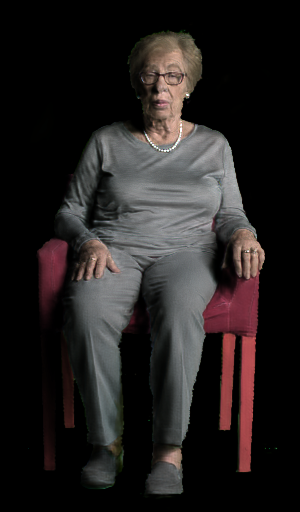}
        \includegraphics[width=\linewidth]{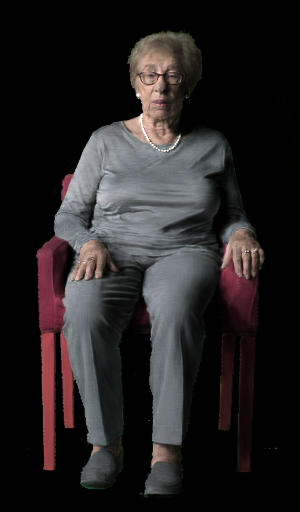}
        \includegraphics[width=\linewidth]{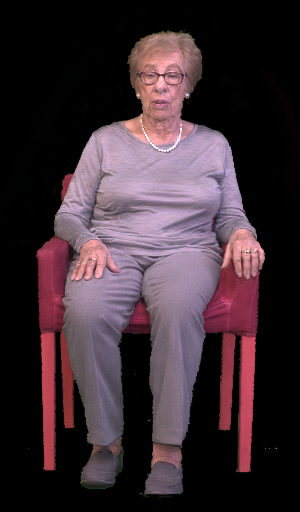}
        \includegraphics[width=\linewidth]{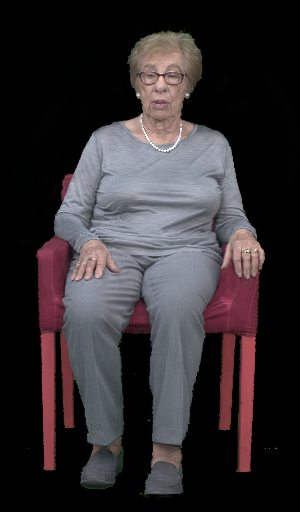}
        % \caption{model B}
    \end{subfigure}\hfill
\caption{\textbf{Relighting results} - Row 1: Input interview videos. Row 2,3: OLAT predictions on two patterns. Row 4,5: Relighting results with two HDRI lighting environments: Grace Cathedral and Pisa Courtyard. See more examples in our video.}
\label{fig:results}
\end{figure*}

We evaluate our technique by relighting several hundred frames of interview footage and comparing to relit images made with that of \cite{Tiancheng19} and \cite{Texler20-CAG}.  We do not have ground truth relighting for each frame of the video to compare to, so we employ a user study to evaluate our method against prior works. Finally, we show how our method is able to realistically relight the dynamic performance of the subject with arbitrary poses and motions.

\subsection{Single Image Portrait Relighting}
We first compared our method with a state-of-the-art lighting estimation and relighting for portrait photos \cite{Tiancheng19}. Their neural network was trained on a dataset of numerous synthetically relit portrait images of 18 individuals from pre-captured OLAT data. From Figure \ref{fig:vsPortrait}, we can see that our method performs much more believable relighting, as the single image portrait relighting result only reproduces the low-frequency components of the novel illumination.  Note also that we cropped our method's result down to just the face to match the output capability of the Single Image Portrait Relighting network, whereas our model is able to relight more of the body as shown in Figure \ref{fig:style}.

\subsection{Relighting as Style Transfer}
We next compared our approach with the state-of-the-art style transfer technique of \cite{Texler20-CAG} that takes several keyframes to use as style and transfer the styles or relighting from those keyframes to the video. As we can see from Figure \ref{fig:style}, the shading on the inner palms of the actor is not supposed to be in shadow, but since the provided keyframes do not cover this pose, \cite{Texler20-CAG} predicts the wrong shading in this area. In contrast, thanks to self-supervised learning, our network is able to recover a more reasonable rendition of the shading one would expect for this pose. For side by side comparison, see our supplementary video.

% \begin{figure}[t]
% \begin{center}
% % \fbox{\rule{0pt}{2in} \rule{0.9\linewidth}{0pt}}
%   \includegraphics[width=\linewidth]{figs/vsstyle.jpg}
% \end{center}
%   \caption{Comparison with Style Transfer based relighting. Left: Style Transfer \cite{Texler20-CAG}, Right: Our result }
% \label{fig:style}
% \end{figure}

\subsection{User Study}
We conducted a user study to evaluate which relighting technique produced preferable results.  We showed users a reference image of the subject under one of the OLAT conditions, and then short video clips of the subject's interview re-lit by that condition using our approach, Single Image Portrait Relighting \cite{Tiancheng19}, and Style Transfer based relighting \cite{Texler20-CAG}.  We finally asked users two questions: 1) Which video clip looks more like the reference image, and 2) Which video clip looks better?  From 61 responses, all users answered both questions with the same answer: 52 chose the video clip rendered with our approach, and 9 chose the video rendered with \cite{Texler20-CAG}, while none chose \cite{Tiancheng19}. This showed a clear preference for our approach.

\subsection{Relighting Dynamic Performance}
We perform relighting for interview footage of three Holocaust survivors. The first survivor was recorded in 2012, while the other two survivors were recorded in 2015. In 2012, the OLAT set consists of 41 patterns while there are 146 patterns used in 2015. Because our method is not restricted to any OLAT patterns, it can generalize to the new setup as long as the diffuse lighting condition from above is guaranteed. For consistency, we choose evenly distributed 41 out of 146 OLAT patterns to train our network.
It is important to note that none of the evaluated interview videos are used to train the neural network. As we can see from Figure \ref{fig:results}, our network is able to predict convincing reflectance fields for novel poses in the interview videos, enabling it to realistically place these interviews in any lighting environment.

\section{Future Work}

In this project, we made use of both the diffusely-lit interview footage and the reflectance field exemplars of each subject, but we only used a single one of the available viewpoints in the data.  It seems possible that even better relighting results could be obtained by leveraging some or all of the views of the subject from the other cameras' positions, even though these other views are also recorded in the same diffuse interview lighting.  The reason is that the multiple viewpoints carry additional information about the subject's three-dimensional shape, and knowing the subject's 3D shape is also useful for predicting their shading and shadowing under new lighting conditions.  For future work, it would be of interest to use the 50 viewpoints available to reconstruct a 3D model such as a Neural Radiance Field \cite{Mildenhall:2020:NRS} for each frame of the interview footage and to leverage these models during training so that the network is better able to learn how shape and the appearance under novel illumination are connected.  However, at this time, such reconstruction techniques might be prohibitively expensive to run on hours of video material.

\section{Conclusion}

In this paper, we presented a deep learning-based video relighting technique that takes diffusely lit video and a set of reflectance field exemplars of the same subject as input.  We designed this technique to work with the data available from the Holocaust survivor interviews recorded in 2014 in the New Dimensions in Testimony project and showed how we can realistically render the Holocaust survivor interview footage in novel lighting conditions.  The technique suggests that this approach could be used to obtain high-quality relighting of new interview footage, assuming that the subjects can also be recorded under a variety of directional lighting conditions in a number of static poses.  This provides the relighting network with subject-specific information for how to relight the video than just the single interview lighting condition alone.

\section{Acknowledgement}
We wish to thank the SHOAH foundation for their effort in making the project New Dimension in Testimony possible and for sharing precious data of the three Holocaust survivors Pinchas Gutter, Aaron Elster, Eva Schloss. We thank Kathleen Haase and Christina Trejo for their coordination. We thank Yajie Zhao and Mingming He for their insightful suggestions. This research was sponsored by the Army Research Office and was accomplished under Cooperative Agreement Number W911NF-20-2-0053, and sponsored by the U.S. Army Research Laboratory (ARL) under contract number W911NF-14-D-0005, the CONIX Research Center, one of six centers in JUMP, a Semiconductor Research Corporation (SRC) program sponsored by DARPA and in part by the ONR YIP grant N00014-17-S-FO14. The views and conclusions contained in this document are those of the authors and should not be interpreted as representing the official policies, either expressed or implied, of the Army Research Office or the U.S. Government. The U.S. Government is authorized to reproduce and distribute.

{\small
\bibliographystyle{ieee_fullname}
\bibliography{egbib}
}

\end{document}